%% file: main.tex
\documentclass[runningheads]{llncs}

\usepackage{eccv}

\usepackage{eccvabbrv}

\usepackage{graphicx}
\usepackage{booktabs}

\usepackage[accsupp]{axessibility}
\usepackage{transparent}

\usepackage{hyperref}

\usepackage{orcidlink}

\graphicspath{{figs/}}

\input{acronyms}

\begin{document}

\title{Warping Earth Observations for better ice labeling in the Marginal Marginal Ice Zone}
\titlerunning{Warping EO for MIZ Labelling}

\author{Tom Kelly \and
Martin S. J. Rogers\orcidlink{0000-0003-0056-2030}}

\authorrunning{T.~Kelly and M.~S.~J.~Rogers}

\institute{British Antarctic Survey, Cambridge \\
\email{\{thokel, marrog\}@bas.ac.uk}}

\maketitle

\input{abstract}

\acresetall

\input{introduction}
\input{related_work}

\input{methods}
\input{experiments}

\input{conclusion}

\clearpage

\bibliographystyle{splncs04}
\bibliography{references}

\end{document}

%% file: acronyms.tex
\usepackage{acro}

\DeclareAcronym{eo}{
  short = EO,
  long  = Earth Observation
}
\DeclareAcronym{s1}{
  short = S1,
  long  = Sentinel-1
}
\DeclareAcronym{sar}{
  short = SAR,
  long  = Synthetic Aperture Radar
}
\DeclareAcronym{modis}{
  short = MODIS,
  long  = Moderate Resolution Imaging Spectroradiometer
}
\DeclareAcronym{miz}{
  short = MIZ,
  long  = Marginal Ice Zone
}
\DeclareAcronym{bacc}{
  short = bAcc,
  long  = Balanced Accuracy
}
\DeclareAcronym{lsvm}{
  short = LSVM,
  long  = Linear Support Vector Machine
}
\DeclareAcronym{gb}{
  short = GB,
  long  = Gradient Boosting
}
\DeclareAcronym{amsr}{
  short = AMSR,
  long  = Advanced Microwave Scanning Radiometer
}

\DeclareAcronym{msd}{
  short = MSD,
  long  = Mean Signed Distance
}

\DeclareAcronym{gfm}{
  short = GFM,
  long  = Geospatial Foundation Model
}

%% file: abstract.tex
\begin{abstract}

Multimodal satellite imagery provides complementary information for Earth Observation, but accurately combining heterogeneous sensors remains challenging in dynamic environments. Fast-changing regions, such as the Antarctic marginal ice zone, cannot fully exploit multimodal information from different satellite sensors because surface features move between image acquisitions. This spatial and temporal mismatch challenges effective \emph{perceptual grounding}, violating the assumption of pixel-level correspondence that underpins most multimodal reasoning and downstream classification pipelines. Antarctic sea ice provides a challenging benchmark due to the rapid, heterogeneous drift of individual ice floes and the differing responses of sea ice to radar, visible and thermal sensing modalities. Accurate, dense supervision of sea ice remains scarce because generating pixel-wise labels requires time-consuming expert interpretation of noisy data, leading to historical reliance on coarse-resolution maritime ice charts for model training. This paper presents a novel architecture based on mutual information warping to align multi-satellite (\acl{s1} and \acl{modis} platforms) multimodal (visible, thermal, radar) satellite scenes. To demonstrate the approach, we introduce a sparse expert-labeled dataset of 2,088 pixel-wise annotations (7,046 expert point classifications) located at the ice-water margin interface across 43 scenes. Our results demonstrate that spatially grounding and aligning modalities prior to segmentation improves classification accuracy, and enables accurate, dense sea ice segmentation from sparse point-wise supervision.

\keywords{Perceptual Grounding \and Multimodal Reasoning \and Evidence Localization \and Sea Ice \and Image Warping \and Earth Observation}
\end{abstract}

%% file: introduction.tex
\begin{figure}[!ht]
    \centering
    \def\svgwidth{\linewidth}
    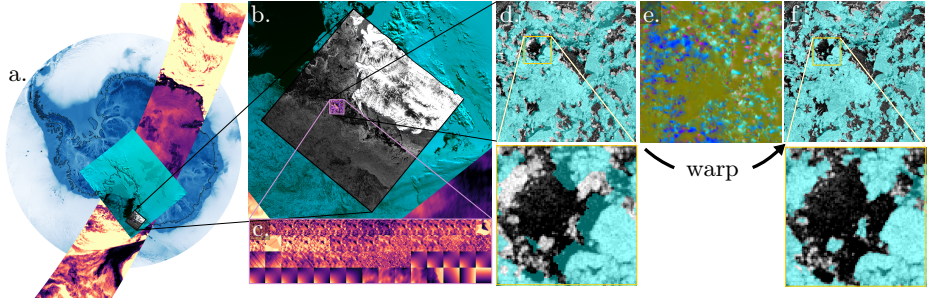
    \caption{Different satellite sources image Antarctica at various locations and scales, taken at similar, but not identical, times (a, b), leading to a lack of alignment on fast-moving objects such as sea ice across modalities (c). We demonstrate that by aligning these modalities with warping, we improve learning results; (d) shows the ice masks from one satellite (MODIS, blue) and a second (S1, gray) before our warping technique, and (f) after. The optimized warp map displaces one set of channels to align to another (e: horizontal displacement (red), vertical (green), magnitude(blue)).}
    \label{fig:teaser}
\end{figure}

\section{Introduction}
Climate change is driving Antarctica towards an increasingly low-ice state, reshaping global ocean circulation, climate, and marine navigation, making accurate and timely sea ice monitoring essential \cite{raphael2025seaiceState, vihma2026antarctic, riihela2021albedo, aksenov2017nav}. Meanwhile, the proliferation of \ac{eo} satellites has created a vast catalog of imagery, spurring rapid advances in multimodal foundation and self-supervised models for downstream forecasting, segmentation, and classification tasks.

Different sensors provide complementary information. ESA's \ac{s1} satellites acquire high-resolution (20-80 meter), cloud-penetrating \ac{sar} imagery, but their backscatter measurements are difficult to interpret, even for experienced sea ice analysts \cite{Najem_incidence}. In contrast, NASA's \ac{modis} satellite captures hyper-spectral imagery with a wide field of view, allowing temporal alignment with \ac{s1}, but at the cost of lower spatial resolution (250-1000 meters).

Despite these advancements, existing pipelines typically assume static scenes, overlooking temporal displacement between acquisitions and sensor misalignment in rapidly evolving environments. This limitation extends beyond sea ice to other dynamic EO applications, including ocean surface processes, cloud systems, floods and natural hazards, e.g., wildfires. Sea ice is particularly dynamic and can drift more than 50 km per day~\cite{farooq2020driftRate, alberello2020drift}, meaning even acquisitions separated by less than an hour can exhibit substantial pixel-wise displacements. While multiple approaches for multimodal sea ice segmentation have been developed, none of these consider the movement of sea ice between acquisitions. This low-level misalignment poses a fundamental problem for multimodal representation learning, preventing accurate cross-modal correspondence and limiting the quality of labeled supervision available for downstream vision tasks.

Our contributions are as follows:
\begin{itemize}
    \item An architecture for warping different satellite modalities using mutual information.
    \item A sparse dataset of 2,088 expert-labeled pins that identify ice in difficult marginal conditions.
    \item A demonstration that warping improves segmentation performance and overall accuracy at boundary locations.
    \item An application to the dense labeling of sea ice.
\end{itemize}

%% file: figs/teaser2.pdf_tex
%% Creator: Inkscape 1.2.2 (b0a8486541, 2022-12-01), www.inkscape.org
%% Accompanies image file 'teaser2.pdf' (pdf, eps, ps)
\begingroup%
  \makeatletter%
  \providecommand\color[2][]{%
    \errmessage{(Inkscape) Color is used for the text in Inkscape, but the package 'color.sty' is not loaded}%
    \renewcommand\color[2][]{}%
  }%
  \providecommand\transparent[1]{%
    \errmessage{(Inkscape) Transparency is used (non-zero) for the text in Inkscape, but the package 'transparent.sty' is not loaded}%
    \renewcommand\transparent[1]{}%
  }%
  \providecommand\rotatebox[2]{#2}%
  \newcommand*\fsize{\dimexpr\f@size pt\relax}%
  \newcommand*\lineheight[1]{\fontsize{\fsize}{#1\fsize}\selectfont}%
  \ifx\svgwidth\undefined%
    \setlength{\unitlength}{598.91742856bp}%
    \ifx\svgscale\undefined%
      \relax%
    \else%
      \setlength{\unitlength}{\unitlength * \real{\svgscale}}%
    \fi%
  \else%
    \setlength{\unitlength}{\svgwidth}%
  \fi%
  \global\let\svgwidth\undefined%
  \global\let\svgscale\undefined%
  \makeatother%
  \begin{picture}(1,0.32026861)%
    \lineheight{1}%
    \setlength\tabcolsep{0pt}%
    \put(0,0){\includegraphics[width=\unitlength,page=1]{teaser2.pdf}}%
    \put(0.02701103,0.23383505){\color[rgb]{0,0,0}\makebox(0,0)[rt]{\lineheight{1.25}\smash{\begin{tabular}[t]{r}a.\end{tabular}}}}%
    \put(0.26862237,0.29772792){\color[rgb]{1,1,1}\makebox(0,0)[lt]{\lineheight{1.25}\smash{\begin{tabular}[t]{l}b.\end{tabular}}}}%
    \put(0,0){\includegraphics[width=\unitlength,page=2]{teaser2.pdf}}%
    \put(0.27040416,0.06592564){\color[rgb]{1,1,1}\makebox(0,0)[lt]{\lineheight{1.25}\smash{\begin{tabular}[t]{l}c.\end{tabular}}}}%
    \put(0,0){\includegraphics[width=\unitlength,page=3]{teaser2.pdf}}%
    \put(0.53818349,0.29772792){\color[rgb]{1,1,1}\makebox(0,0)[lt]{\lineheight{1.25}\smash{\begin{tabular}[t]{l}d.\end{tabular}}}}%
    \put(0,0){\includegraphics[width=\unitlength,page=4]{teaser2.pdf}}%
    \put(0.69337526,0.29790161){\color[rgb]{1,1,1}\makebox(0,0)[lt]{\lineheight{1.25}\smash{\begin{tabular}[t]{l}e.\end{tabular}}}}%
    \put(0,0){\includegraphics[width=\unitlength,page=5]{teaser2.pdf}}%
    \put(0.85303314,0.29750186){\color[rgb]{1,1,1}\makebox(0,0)[lt]{\lineheight{1.25}\smash{\begin{tabular}[t]{l}f.\end{tabular}}}}%
    \put(0,0){\includegraphics[width=\unitlength,page=6]{teaser2.pdf}}%
    \put(0.76877768,0.13165485){\color[rgb]{0,0,0}\makebox(0,0)[t]{\lineheight{1.25}\smash{\begin{tabular}[t]{c}warp\end{tabular}}}}%
    \put(0,0){\includegraphics[width=\unitlength,page=7]{teaser2.pdf}}%
  \end{picture}%
\endgroup%

%% file: related_work.tex
\section{Related Work}

\textbf{Sea Ice Classification and Segmentation.} Producing dense sea ice classifications has traditionally relied heavily on \ac{sar} imagery and manually interpreted ice charts \cite{stokholm2024autoice, wulf2026sar, stokholm2023ai4seaice, boulze2020saronly, pires2023saratrous}. Early automated approaches remain challenging because ambiguous backscatter signals and acquisition geometry can cause overlapping backscatter signatures, making ice-water discrimination challenging \cite{Najem_incidence}. Recent work demonstrates that combining multiple SAR polarizations with visible \cite{rogers2024sea, konig2021sarvis}, thermal \cite{khachatrian2026, radhakrishnan2024autoice}, passive microwave \cite{wulf2024sarpwm}, and climatic variables \cite{de2021sarwind} significantly improves class separation. To scale these analyses, operational pipelines have transitioned from classical classifiers to deep convolutional architectures, predominantly employing extended-receptive-field UNet \cite{ronneberger2015u} variants \cite{radhakrishnan2024autoice, stokholm2023ai4seaice} or more recently transformer architectures \cite{alkaee2025iceBench}. Modalities have been fused via low-level input concatenation \cite{stokholm2024autoice, de2021sarwind, wulf2024sarpwm} or higher-level deep-layer fusion \cite{rogers2024sea}, enabling models to learn complementary representations across sensors. However, the scarcity of high-quality sea ice labels remains a major limitation. Producing accurate annotations is time-consuming and dependent upon a small number of experienced sea ice analysts to interpret noisy \ac{sar} imagery. Existing labeled datasets lack scale~\cite{li2024aiceReview}, only sample non-ice edges~\cite{khachatrian2026}, utilize oversimplified thresholding techniques \cite{rogers2024sea}, or provide inaccurate polygonal segmentation over the fractal-like ice-water boundary \cite{wulf2026sar, stokholm2024autoice}.
While curating expertly labeled, sparse data would reduce the ingestion of contaminated pixels \cite{heffring2026, khachatrian2026}, these pixel-wise labeled datasets remain rare. Furthermore, despite advances in architecture design and supervision strategies, these methods generally assume that multimodal observations are spatially aligned.

\textbf{Multimodal Foundation Models for EO.} There has been a rapid expansion in foundation models adapted for \ac{eo}. These self-supervised \acp{gfm} leverage vast, unlabeled multimodal catalogs for tasks ranging from forecasting to segmentation, employing explicit 3D spatiotemporal patch embeddings \cite{prithvi2024} and coordinate encoders to encapsulate topological context \cite{satclip2023}. State-of-the-art architectures target the complete unification of disparate sensor modalities through multi-granularity contrastive learning \cite{skysense2023} and wavelength-conditioned dynamic hypernetworks \cite{dofa2024}, enabling the seamless imputation of missing bands via generative architectures like EDCGAN \cite{chouhan202x}. However, zero-shot deployment of mid-latitude GFMs into the polar cryosphere reveals a severe domain gap, strictly mandating partial fine-tuning of encoder layers via few-shot learning to exceed standard convolutional baselines \cite{cryobench2026}. Crucially, while these approaches fuse multimodal data to improve classification, they assume static scenes between acquisitions \cite{prithvi2024, satclip2023, skysense2023}, generally aligning pixels based on spatial coordinates alone and disregarding the sub-daily motion of the underlying surface.

\textbf{Image Alignment and Drift Compensation.}
Estimating dense sea ice motion fields has long been studied for forecasting and drift analysis. Early approaches relied on classical optical flow and feature tracking between consecutive SAR acquisitions \cite{muckenhuber2016icedriftorb, howell2022iceDtift, korosov2017iceDriftFeat}, but classical motion estimation heuristics struggle under large non-rigid deformation, ambiguous backscatter, and image noise. More recently, unsupervised deep learning approaches have substantially improved dense motion estimation by learning non-linear displacement fields directly from image pairs, offering greater robustness to changing image conditions to extract dense, sub-kilometer drift fields \cite{martin2025, uusinoka2025deepIceDrift, gao2025iceDriftDL}. However, these approaches estimate motion within a single sensing modality and are not designed to establish dense correspondence across heterogeneous satellite observations.
Dense multimodal registration has instead been explored for training deformable registration networks to learn correspondences between medical images \cite{balakrishnan2019voxelmorph, deng2024Voxdev, lee2023seq2morphVoxDev, feenstra2024deformableVoxDev}. Transformer-based architectures have further improved non-rigid alignment accuracy by tracking global structural dependencies \cite{chen2022transmorph, samir2025}. Although these methods demonstrate the feasibility of cross-modal registration, they assume comparatively controlled imaging geometries and static anatomy assumptions that do not hold for rapidly evolving \ac{eo} scenes. To our knowledge, unsupervised deformable registration to align EO imagery across time, image spatial resolutions, and modalities remains underexplored, where both modality differences and physical surface motion must be jointly addressed.

%% file: methods.tex
\section{Method}

\subsection{Mutual Information Warping Architecture}
\label{sec:warp_arch}

\begin{figure}[!ht]
    \centering
    \def\svgwidth{0.8\linewidth}
    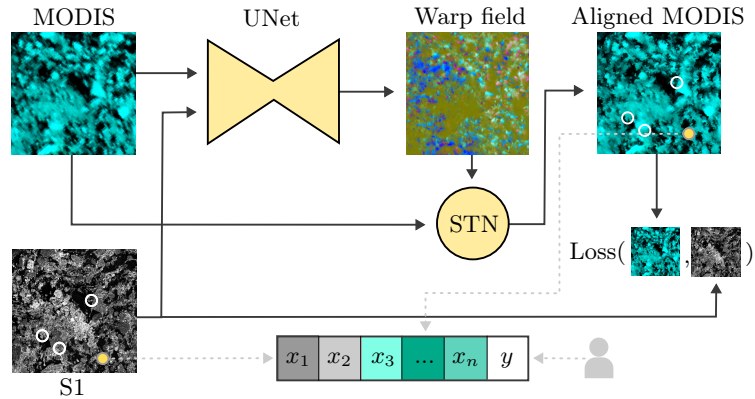
    \caption{In the first stage (solid arrows), we warp the \ac{modis} channels to match the \ac{s1} channels using a warp field created by a UNet. This field is applied to the \ac{modis} channels to create an aligned \ac{modis} image. This optimization takes place on a loss, $ \mathcal{L}$. In the second stage the warped dataset is built from the pixel feature vectors ($x_n$) with the expert's ice/water labels (circles, $y$).}
    \label{fig:arch}
\end{figure}

We observe that ice is more easily identified in \ac{modis} imagery, but it is not well aligned with the higher-resolution \ac{s1} channels at the pixel level. Our approach is to warp the lower-resolution \ac{modis} to match the \ac{s1} using a spatial transformer network~\cite{jaderberg2015spatial} to create a per-pixel warp (displacement) field which translates each \ac{modis} channel to align with the \ac{s1}, simultaneously increasing the native spatial resolution of \ac{modis} to match \ac{s1}. This approach is applied in medical imaging to align deformable organs~\cite{balakrishnan2019voxelmorph} between scans with the same modality. However, the larger differences in appearance between satellite modalities cause Mean Squared Error (MSE), Laplacian, or cross-correlation losses to fail.

In the visible \ac{modis} channels, the transition from ice to water typically appears as a transition from white to black. However, in \ac{sar} images, the high-low backscatter transition may be reversed (low-high) due to low satellite incidence angles or smooth melting ice. Given these challenges, we found that a robust alignment is provided by maximizing the Local Mutual Information (LMI) between the \ac{sar} backscatter and the \ac{modis} thermal/optical reflectance. LMI can identify misaligned spatial features and pull them together in a wide variety of geographic features. In addition, the physics of ice is a well-known prior, our objective function also includes informed regularization to cap the speed and compressibility of sea ice. Together with standard terms such as smoothness and a zero-movement prior, these allow us to find a suitable warp field between modalities. Therefore, our objective function is a composite loss to align the distinct sensor distributions while enforcing physically plausible deformations:

\begin{equation}
    \mathcal{L} = \mathcal{L}_{LMI} + \lambda_{jac} \mathcal{L}_{jac} + \lambda_{smooth} \mathcal{L}_{smooth} + \lambda_{cap} \mathcal{L}_{cap} + \lambda_{zero} \mathcal{L}_{zero}
\end{equation}

Minimizing $\mathcal{L}_{LMI}$ maximizes the statistical dependence between the normalized \ac{sar} backscatter and \ac{modis} reflectance. By computing probability distributions over a local spatial window, it effectively aligns disparate modalities where global intensity mappings fail, bridging non-linear radiometric differences between sensors.

\begin{equation}
    \mathcal{L}_{LMI} = - \frac{1}{|\Omega|} \sum_{x \in \Omega} \left( H(X_x) + H(Y_x) - H(X_x, Y_x) \right)
\end{equation}

where $H$ denotes the Shannon entropy computed via soft Gaussian binning over a local neighborhood at pixel $x$, $\Omega$ is the spatial domain ($31 \times 31$ pixels), and  $X$ and $Y$ are the images to align (\ac{sar} and \ac{modis} channels). The Jacobian Loss ($\mathcal{L}_{jac}$) encourages physical flow by penalizing local expansion or compression to enforce area preservation. This acts as a physical constraint on sea ice drift, helping the modeled deformation remain incompressible over the image.

\begin{equation}
    \mathcal{L}_{jac} = \frac{1}{|\Omega|} \sum_{x \in \Omega} (|J_\phi(x)| - 1)^2
\end{equation}

where $J_\phi$ is the Jacobian determinant of the spatial transformation $\phi(x) = x + \mathbf{u}(x)$. The Smoothness Loss ($\mathcal{L}_{smooth}$) enforces spatial coherence in the continuous warp field, preventing non-physical tearing or folding by penalizing large local gradients in the displacement vectors $\mathbf{u}$:
\begin{equation}
    \mathcal{L}_{smooth} = \frac{1}{|\Omega|} \sum_{x \in \Omega} \left( \|\nabla_x \mathbf{u}(x)\|_1 + \|\nabla_y \mathbf{u}(x)\|_1 \right)
\end{equation}

The Magnitude Capping Loss ($\mathcal{L}_{cap}$) restricts the maximum permitted displacement to reflect physical boundaries of sea ice drift during the temporal gap. A quadratic penalty is applied exclusively to vectors exceeding a predefined pixel threshold $u_{max}$:
\begin{equation}
    \mathcal{L}_{cap} = \frac{1}{|\Omega|} \sum_{x \in \Omega} \max(0, \|\mathbf{u}(x)\|_2 - u_{max})^2
\end{equation}

Finally, the Zero-Displacement Loss ($\mathcal{L}_{zero}$) is an $L_1$ sparsity regularizer. It encourages the network to default to zero displacement in regions of high ambiguity (such as featureless open water), preventing spurious deformations driven by noise:
\begin{equation}
    \mathcal{L}_{zero} = \frac{1}{|\Omega|} \sum_{x \in \Omega} \|\mathbf{u}(x)\|_1
\end{equation}

This loss drives the weights of a UNet \cite{ronneberger2015u}, whose multi-scale resolution combines global smoothness and local detail. The symmetric UNet encoder takes 512px inputs and contains four layers, with a 32-channel bottleneck.

Optimization takes place over the combined loss using the parameters $\lambda_{jac}=1 \times 10^{-3}$, $\lambda_{smooth} = 5.72 \times 10^{-2}$, $\lambda_{cap} = 9.872$, and $\lambda_{zero} = 6.946 \times 10^{-7}$. We compute the $\mathcal{L}_{LMI}$ between the \ac{s1} HH polarization and \ac{modis} channel 2. These parameters were found over a sweep of 152 parameter sets. The UNet parameters are optimized directly using Adam~\cite{kingma2014adam} with $\beta_{1} = 0.9, \beta_{2} = 0.999$, and a learning rate of $1 \times 10^{-4}$ for 500 iterations with a cosine annealing learning rate scheduler.

\subsection{Data and Sparse Label Collection}

To validate our approach, we collected a sparse dataset of high-precision pixel labels. It comprises multispectral imagery from the \ac{modis} Aqua and Terra satellites and \ac{sar} scenes from Sentinel-1A and Sentinel-1B. Following standard convention \cite{gorelick2017google}, the \ac{sar} data is scaled to decibels. We utilize the HH, HV polarizations from \ac{s1} (2 channels), and all visible and thermal channels from \ac{modis} (38 channels). Auxiliary channels include low-resolution passive microwave \ac{amsr} (16 channels) and static topography features (2 channels). Locations were sampled from the Antarctic \ac{miz}, and the time gap between \ac{s1} and \ac{modis} acquisitions was limited to 1 hour to bound the maximum ice drift between images.

\begin{figure}[!ht]
    \centering
    \def\svgwidth{\linewidth}
    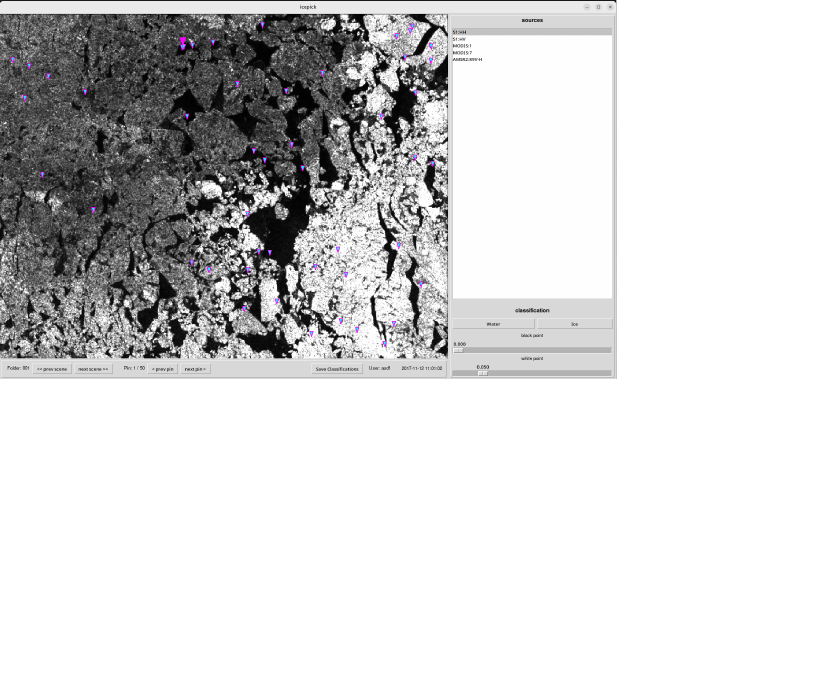
    \caption{The user interface (above) and the \ac{modis} channel 1 used to distribute pins over the marginal ice (below: left shows same scene as above). The users can zoom and pan with the mouse.}
    \label{fig:pin_tool}
\end{figure}

 To address the limitations in the quality of pre-existing sea ice labels, the dataset emphasizes quality over quantity: four ice experts labeled up to 2,088 sparse pins each, distributed over the most challenging ice-water boundary pixels. These locations were selected by normally sampling \ac{modis} channel 1 around the interface (as the ice/water transition is typically light/dark in the visible channels), generating 7,046 total classifications across 43 distinct areas~\footnote{\url{https://github.com/twak/marginal_marginal}}. The experts labeled the pixels of the high-resolution \ac{s1} channel with access to coincident \ac{modis} and \ac{amsr} to inform their decisions. As shown in Figure~\ref{fig:pin_tool}, the experts used a custom interface to label individual pixels.

\begin{figure}[!ht]
    \centering
    \def\svgwidth{\linewidth}
    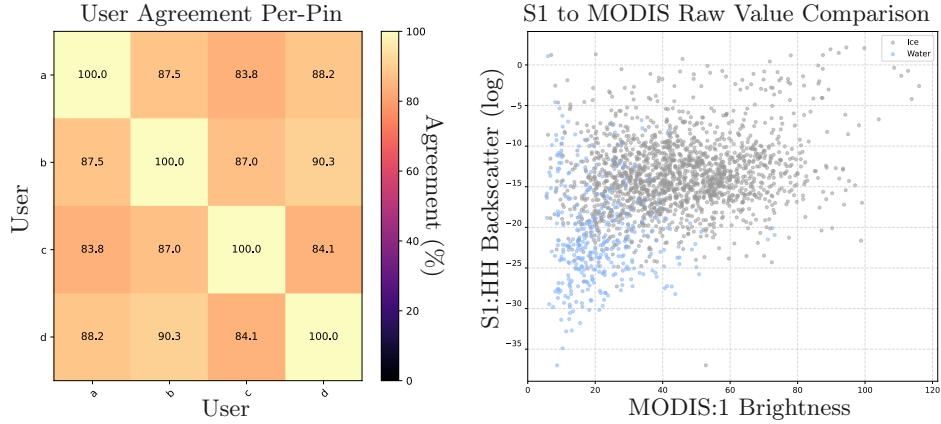
    \caption{Dataset statistics demonstrating strong agreement among experts (left) and non-trivial overlap between classifications and two representative modalities (right).}
    \label{fig:pin_stats}
\end{figure}

 This dataset results in binary labels with point-wise classifications and corresponding satellite features per point. While not all experts labeled every pin, we observed strong agreement (Figure~\ref{fig:pin_stats}). The challenging task ensured some disagreement; an oracle achieves a \ac{bacc} of 0.91. The dataset is split into a 60/20/20 train/val/test configuration, with pins in separate geographic chips in different partitions to allow dense UNet labeling.

%% file: figs/arch.pdf_tex
%% Creator: Inkscape 1.2.2 (b0a8486541, 2022-12-01), www.inkscape.org
%% Accompanies image file 'arch.pdf' (pdf, eps, ps)
\begingroup%
  \makeatletter%
  \providecommand\color[2][]{%
    \errmessage{(Inkscape) Color is used for the text in Inkscape, but the package 'color.sty' is not loaded}%
    \renewcommand\color[2][]{}%
  }%
  \providecommand\transparent[1]{%
    \errmessage{(Inkscape) Transparency is used (non-zero) for the text in Inkscape, but the package 'transparent.sty' is not loaded}%
    \renewcommand\transparent[1]{}%
  }%
  \providecommand\rotatebox[2]{#2}%
  \newcommand*\fsize{\dimexpr\f@size pt\relax}%
  \newcommand*\lineheight[1]{\fontsize{\fsize}{#1\fsize}\selectfont}%
  \ifx\svgwidth\undefined%
    \setlength{\unitlength}{403.27808771bp}%
    \ifx\svgscale\undefined%
      \relax%
    \else%
      \setlength{\unitlength}{\unitlength * \real{\svgscale}}%
    \fi%
  \else%
    \setlength{\unitlength}{\svgwidth}%
  \fi%
  \global\let\svgwidth\undefined%
  \global\let\svgscale\undefined%
  \makeatother%
  \begin{picture}(1,0.51858012)%
    \lineheight{1}%
    \setlength\tabcolsep{0pt}%
    \put(0,0){\includegraphics[width=\unitlength,page=1]{arch.pdf}}%
    \put(0.35250652,0.49902814){\color[rgb]{0,0,0}\makebox(0,0)[t]{\lineheight{1.25}\smash{\begin{tabular}[t]{c}UNet\end{tabular}}}}%
    \put(0.91122974,0.17825647){\color[rgb]{0,0,0}\makebox(0,0)[lt]{\lineheight{1.25}\smash{\begin{tabular}[t]{l},\end{tabular}}}}%
    \put(0.08477303,0.50108184){\color[rgb]{0,0,0}\makebox(0,0)[t]{\lineheight{1.25}\smash{\begin{tabular}[t]{c}MODIS\end{tabular}}}}%
    \put(0.08251638,0.00022011){\color[rgb]{0,0,0}\makebox(0,0)[t]{\lineheight{1.25}\smash{\begin{tabular}[t]{c}S1\end{tabular}}}}%
    \put(0.62169858,0.50173808){\color[rgb]{0,0,0}\makebox(0,0)[t]{\lineheight{1.25}\smash{\begin{tabular}[t]{c}Warp field\end{tabular}}}}%
    \put(0.87634489,0.50179342){\color[rgb]{0,0,0}\makebox(0,0)[t]{\lineheight{1.25}\smash{\begin{tabular}[t]{c}Aligned MODIS\end{tabular}}}}%
    \put(0,0){\includegraphics[width=\unitlength,page=2]{arch.pdf}}%
    \put(0.83279379,0.18453041){\color[rgb]{0,0,0}\makebox(0,0)[rt]{\lineheight{1.25}\smash{\begin{tabular}[t]{r}Loss(\end{tabular}}}}%
    \put(0.9942759,0.18295174){\color[rgb]{0,0,0}\makebox(0,0)[lt]{\lineheight{1.25}\smash{\begin{tabular}[t]{l})\end{tabular}}}}%
    \put(0,0){\includegraphics[width=\unitlength,page=3]{arch.pdf}}%
    \put(0.67154636,0.04253832){\color[rgb]{0,0,0}\makebox(0,0)[t]{\lineheight{1.25}\smash{\begin{tabular}[t]{c}$y$\end{tabular}}}}%
    \put(0,0){\includegraphics[width=\unitlength,page=4]{arch.pdf}}%
    \put(0.62710612,0.22084764){\color[rgb]{0,0,0}\makebox(0,0)[t]{\lineheight{1.25}\smash{\begin{tabular}[t]{c}STN\end{tabular}}}}%
    \put(0,0){\includegraphics[width=\unitlength,page=5]{arch.pdf}}%
    \put(0.38886316,0.04253832){\color[rgb]{0,0,0}\makebox(0,0)[t]{\lineheight{1.25}\smash{\begin{tabular}[t]{c}$x_1$\end{tabular}}}}%
    \put(0.44465609,0.04253832){\color[rgb]{0,0,0}\makebox(0,0)[t]{\lineheight{1.25}\smash{\begin{tabular}[t]{c}$x_2$\end{tabular}}}}%
    \put(0.50416811,0.04253832){\color[rgb]{0,0,0}\makebox(0,0)[t]{\lineheight{1.25}\smash{\begin{tabular}[t]{c}$x_3$\end{tabular}}}}%
    \put(0.55996082,0.04253832){\color[rgb]{0,0,0}\makebox(0,0)[t]{\lineheight{1.25}\smash{\begin{tabular}[t]{c}...\end{tabular}}}}%
    \put(0.61575359,0.04253832){\color[rgb]{0,0,0}\makebox(0,0)[t]{\lineheight{1.25}\smash{\begin{tabular}[t]{c}$x_n$\end{tabular}}}}%
    \put(0,0){\includegraphics[width=\unitlength,page=6]{arch.pdf}}%
  \end{picture}%
\endgroup%

%% file: figs/pin_tool.pdf_tex
%% Creator: Inkscape 1.2.2 (b0a8486541, 2022-12-01), www.inkscape.org
%% Accompanies image file 'pin_tool.pdf' (pdf, eps, ps)
\begingroup%
  \makeatletter%
  \providecommand\color[2][]{%
    \errmessage{(Inkscape) Color is used for the text in Inkscape, but the package 'color.sty' is not loaded}%
    \renewcommand\color[2][]{}%
  }%
  \providecommand\transparent[1]{%
    \errmessage{(Inkscape) Transparency is used (non-zero) for the text in Inkscape, but the package 'transparent.sty' is not loaded}%
    \renewcommand\transparent[1]{}%
  }%
  \providecommand\rotatebox[2]{#2}%
  \newcommand*\fsize{\dimexpr\f@size pt\relax}%
  \newcommand*\lineheight[1]{\fontsize{\fsize}{#1\fsize}\selectfont}%
  \ifx\svgwidth\undefined%
    \setlength{\unitlength}{390.62235675bp}%
    \ifx\svgscale\undefined%
      \relax%
    \else%
      \setlength{\unitlength}{\unitlength * \real{\svgscale}}%
    \fi%
  \else%
    \setlength{\unitlength}{\svgwidth}%
  \fi%
  \global\let\svgwidth\undefined%
  \global\let\svgscale\undefined%
  \makeatother%
  \begin{picture}(1,0.84821831)%
    \lineheight{1}%
    \setlength\tabcolsep{0pt}%
    \put(0,0){\includegraphics[width=\unitlength,page=1]{pin_tool.pdf}}%
    \put(0.77036277,0.77594791){\color[rgb]{0,0,0}\makebox(0,0)[lt]{\lineheight{1.25}\smash{\begin{tabular}[t]{l}channels\end{tabular}}}}%
    \put(0.76409534,0.47111443){\color[rgb]{0,0,0}\makebox(0,0)[lt]{\lineheight{1.25}\smash{\begin{tabular}[t]{l}classification buttons\end{tabular}}}}%
    \put(0.76444752,0.40359975){\color[rgb]{0,0,0}\makebox(0,0)[lt]{\lineheight{1.25}\smash{\begin{tabular}[t]{l}brightness controls\end{tabular}}}}%
    \put(0.12310695,0.58685006){\color[rgb]{0,1,1}\makebox(0,0)[lt]{\lineheight{1.25}\smash{\begin{tabular}[t]{l}pins\end{tabular}}}}%
    \put(0,0){\includegraphics[width=\unitlength,page=2]{pin_tool.pdf}}%
  \end{picture}%
\endgroup%

%% file: figs/pin_stats_small.pdf_tex
%% Creator: Inkscape 1.2.2 (b0a8486541, 2022-12-01), www.inkscape.org
%% Accompanies image file 'pin_stats_small.pdf' (pdf, eps, ps)
\begingroup%
  \makeatletter%
  \providecommand\color[2][]{%
    \errmessage{(Inkscape) Color is used for the text in Inkscape, but the package 'color.sty' is not loaded}%
    \renewcommand\color[2][]{}%
  }%
  \providecommand\transparent[1]{%
    \errmessage{(Inkscape) Transparency is used (non-zero) for the text in Inkscape, but the package 'transparent.sty' is not loaded}%
    \renewcommand\transparent[1]{}%
  }%
  \providecommand\rotatebox[2]{#2}%
  \newcommand*\fsize{\dimexpr\f@size pt\relax}%
  \newcommand*\lineheight[1]{\fontsize{\fsize}{#1\fsize}\selectfont}%
  \ifx\svgwidth\undefined%
    \setlength{\unitlength}{890.92559814bp}%
    \ifx\svgscale\undefined%
      \relax%
    \else%
      \setlength{\unitlength}{\unitlength * \real{\svgscale}}%
    \fi%
  \else%
    \setlength{\unitlength}{\svgwidth}%
  \fi%
  \global\let\svgwidth\undefined%
  \global\let\svgscale\undefined%
  \makeatother%
  \begin{picture}(1,0.44281018)%
    \lineheight{1}%
    \setlength\tabcolsep{0pt}%
    \put(0,0){\includegraphics[width=\unitlength,page=1]{pin_stats_small.pdf}}%
    \put(0.78321989,0.00386273){\color[rgb]{0.10196078,0.10196078,0.10196078}\makebox(0,0)[t]{\lineheight{1.25}\smash{\begin{tabular}[t]{c}MODIS:1 Brightness\end{tabular}}}}%
    \put(0.52144722,0.23826176){\color[rgb]{0.10196078,0.10196078,0.10196078}\rotatebox{90}{\makebox(0,0)[t]{\lineheight{1.25}\smash{\begin{tabular}[t]{c}S1:HH Backscatter (log)\end{tabular}}}}}%
    \put(0.44360549,0.24094167){\color[rgb]{0.10196078,0.10196078,0.10196078}\rotatebox{-90}{\makebox(0,0)[t]{\lineheight{1.25}\smash{\begin{tabular}[t]{c}Agreement (\%)\end{tabular}}}}}%
    \put(0.01149163,0.22671152){\color[rgb]{0.10196078,0.10196078,0.10196078}\rotatebox{90}{\makebox(0,0)[t]{\lineheight{1.25}\smash{\begin{tabular}[t]{c}User\end{tabular}}}}}%
    \put(0.22607633,0.00226469){\color[rgb]{0.10196078,0.10196078,0.10196078}\makebox(0,0)[t]{\lineheight{1.25}\smash{\begin{tabular}[t]{c}User\end{tabular}}}}%
    \put(0.21062159,0.43057819){\color[rgb]{0.10196078,0.10196078,0.10196078}\makebox(0,0)[t]{\lineheight{1.25}\smash{\begin{tabular}[t]{c}User Agreement Per-Pin\end{tabular}}}}%
    \put(0.76959085,0.43057819){\color[rgb]{0.10196078,0.10196078,0.10196078}\makebox(0,0)[t]{\lineheight{1.25}\smash{\begin{tabular}[t]{c}S1 to MODIS  Raw Value Comparison\end{tabular}}}}%
  \end{picture}%
\endgroup%

%% file: experiments.tex
\section{Experiments}

\begin{table}
\centering
\caption{Top 4 features (univariate \ac{bacc}) for selected \ac{s1} and \ac{amsr} channels.}
\label{tab:base_feature_importance}
\input{base_feature_importance_table}
\end{table}

\subsection{Evaluation Metrics}
Given the complex nature of the \ac{miz}, regular accuracy is often a poor metric due to class imbalances within our dataset. Even though the pixel-wise annotations were positioned at the dynamic ice-water interfaces, inherent imbalances persist in the labeled samples. To provide a fair evaluation that penalizes models exhibiting a bias toward the majority class, we primarily report \ac{bacc} across our experiments. \ac{bacc} normalizes true positive and true negative rates by the number of true instances in each class, ensuring that performance improvements reflect genuine enhancements in distinguishing both ice and water.

\subsection{Individual Feature Importance}
To assess the discriminative power of the available features prior to any warping, we first analyze the individual feature importance using \ac{lsvm} and Histogram \ac{gb} classifiers. Table \ref{tab:base_feature_importance} demonstrates the univariate \ac{bacc} for the high-resolution \ac{s1} (\ac{sar}) and \ac{amsr} (microwave) features. These bands align with pre-existing literature, providing foundational discriminative capability for ice classification.

By comparing the predictive utility of regular \ac{modis} features against their warped counterparts, we observe a significant redistribution of importance when the modalities are physically aligned. Table \ref{tab:modis_feature_importance} demonstrates this effect for the warped \ac{modis} features.

\begin{table}[htpb]
\centering
\caption{Top 15 \ac{modis} Features (Univariate \ac{bacc}) comparing performance with and without warping. Best feature per model in bold.}
\label{tab:modis_feature_importance}
\input{modis_feature_importance_table}
\end{table}

\subsection{Multivariate Performance}

We evaluated performance with and without warping on pairs of channels, as well as all available channels. Learning from pairs of channels showed an uplift over individual high-resolution channels and the majority of channel combinations overall. We observed maximal performance on all 56 channels, evaluating several standard single-pixel segmentation models to quantify performance with and without our warping approach. The unwarped models were trained and tested on the raw, unwarped features. The warped models were trained and tested on the non-\ac{modis} channels and the warped \ac{modis} data. Table \ref{tab:results} summarizes the \ac{bacc} achieved by each model. The results indicate a performance uplift across most models when the modalities are aligned via warping. Notably, the three most performant models all benefited from warping.

\begin{table}[htpb]
\centering
\caption{Comparison of \ac{bacc} for classification models with and without warping, including `All Ice' majority class and human-labeled `Oracle' baselines.}
\label{tab:results}
\input{model_results_table}
\end{table}

\subsection{Larger Spatial Contexts}

Ice labeling in satellite scenes is inherently a spatial (2D) problem. The expectation is that the pixels surrounding a pin provide valuable context. Here we present the results of segmentation using different context-aware techniques. Given that \ac{lsvm} yielded strong initial results, an attractive approach is to expand the feature window provided to traditional ML techniques. However, for both $3 \times 3$ and $5 \times 5$ pixel context windows, the warped \ac{lsvm} \ac{bacc} dropped to 0.8648 and 0.8276, respectively.

A traditional candidate for spatial segmentation is the UNet architecture \cite{ronneberger2015u}. From two basic architectures, a sweep of 256 experiments explored the hyperparameter space for both warped and unwarped data, identifying an optimal model for the unwarped (model A) and warped data (model B). Table~\ref{tab:unet_bAcc} presents the accuracies of these networks. Both networks performed better on the warped data; however, providing a spatial context of either $12 \times 12$ or $20 \times 20$ pixels yielded very limited improvements over a single-pixel LSVM.

\begin{table}[htpb]
\centering
\caption{The \ac{bacc} for the best architectures on both the unwarped (A) and warped (B) data.}
\label{tab:unet_bAcc}

\begin{tabular}{|l|c|c|c|}
\hline
architecture & \begin{tabular}[c]{@{}c@{}}receptive field\\ pixels\end{tabular} & \begin{tabular}[c]{@{}c@{}}\ac{bacc}\\ Unwarped \end{tabular} & \begin{tabular}[c]{@{}c@{}}\ac{bacc}\\ Warped\end{tabular} \\ \hline
A & 12 & 0.8688 & \textbf{0.87319} \\ \cline{1-1}
B & 20 & 0.86106 & {\underline{ \textbf{0.88364}}} \\ \hline
\end{tabular}
\end{table}

\newpage
\subsection{Dense Labeling}
\begin{figure}[!ht]
    \centering
    \def\svgwidth{\linewidth}
    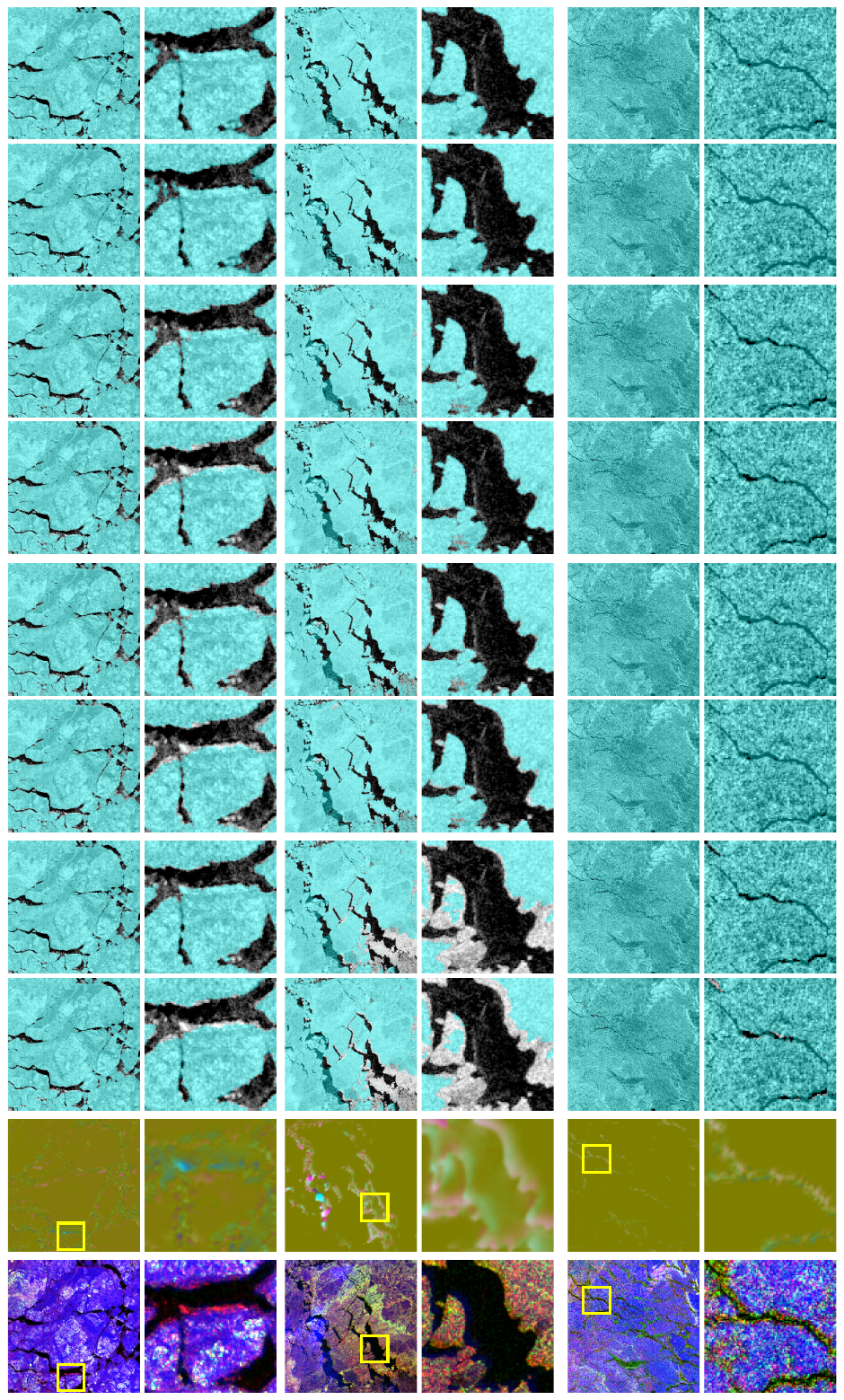
    \caption{Input RGB shows S1:HH, S1:HV, and MODIS channels. The label results show S1:HH in gray and the segmentation in cyan. M:1 LSVM is the first MODIS channel only. The other LSVM and GB columns are created with all channels.}
    \label{fig:egs1}
\end{figure}

\begin{figure}[!ht]
    \centering
    \def\svgwidth{\linewidth}
    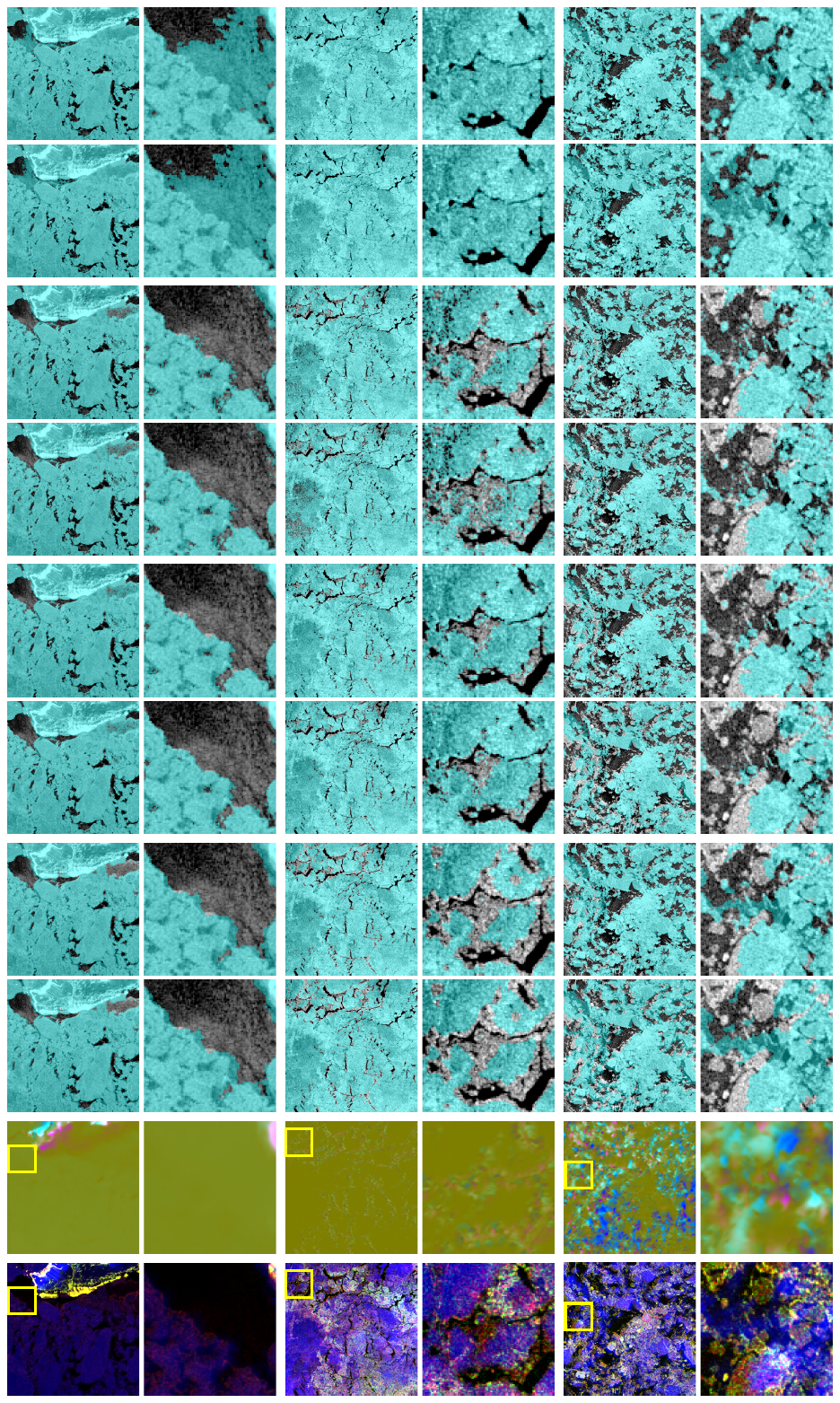
    \caption{As Table~\ref{fig:egs1}, randomly selected chips from the test partition, with randomly (yellow boxes) selected zoom locations over mixed ice/water. The CNN uses the found configuration of base channels and topography.}
    \label{fig:egs2}
\end{figure}

A practical application of classifying the single pixel pins is dense labeling. Figures~\ref{fig:egs1} and~\ref{fig:egs2} illustrate the visual impact of warping on single-channel, single-pixel, and UNet classifications for a sample scene from the test partition. We observe that models utilizing the full suite of aligned channels—particularly the CNN architectures—produce substantially more coherent and physically plausible ice margins.

\subsection{Warp Field Ablations}

To identify the optimal warping hyperparameters (Section~\ref{sec:warp_arch}) given limited training data points, we fit an \ac{lsvm} to the validation partition and calculated the Mean Geometric Margin (Mean Signed Distance) from each sample to the learned decision boundary. A higher mean geometric margin implies that semantically corresponding structures across diverse modalities were more successfully aligned by the warp parameters. Through a Bayesian sweep of 152 parameter configurations, we identified the aforementioned hyperparameters and found that optimizing the mutual information strictly between the S1:HH polarization and MODIS channel 2 produced the most distinct ice/water class separation (Table~\ref{tab:channel_groups}). Formulations that used the Laplacian or MSE loss in place of LMI consistently yielded lower mean geometric margins.

\begin{table}[h]
\centering
\begin{tabular}{|l|c|}
\hline
Channel Groups & \ac{msd} \\
\hline
S1:HH, S1:HV $\leftrightarrow$ MODIS:1, MODIS:2 &  0.8267 \\
S1:HH $\leftrightarrow$ MODIS:1 & 0.8254 \\
\textbf{S1:HH $\leftrightarrow$ MODIS:2} &  \textbf{0.8359} \\
S1:HV $\leftrightarrow$ MODIS:1 & 0.8014 \\
S1:HV $\leftrightarrow$ MODIS:2 &  0.8348 \\
S1:HH $\leftrightarrow$ MODIS:1 \& & 0.7815 \\
S1:HH $\leftrightarrow$ MODIS:2 \& &\\
S1:HV $\leftrightarrow$ MODIS:1 \& &\\
S1:HV $\leftrightarrow$ MODIS:2 &  \\
\hline
\end{tabular}
\caption{Effect of Channel Groups on \ac{msd}, where $\leftrightarrow$ is the LMI, from which we take the mean difference between the groups (separated by a \&). }
\label{tab:channel_groups}
\end{table}

%% file: base_feature_importance_table.tex
\begin{tabular}{|l|c|c|}
\hline
 & \ac{lsvm} & \ac{gb} \\
 channel & \ac{bacc} & \ac{bacc} \\
\hline
s1:hh & 0.7741 & \underline{\textbf{0.7925}} \\
s1:hv & \textbf{0.7765} & 0.7620 \\
amsr:btemp\_89.0ah & \textbf{0.5837} & 0.5746 \\
amsr:btemp\_89.0bh & \textbf{0.5866} & 0.5691 \\
\hline
\end{tabular}

%% file: modis_feature_importance_table.tex
\begin{tabular}{|l|c|c|c|c|}
\hline
 & \ac{lsvm} & \ac{lsvm}& \ac{gb} & \ac{gb} \\
channel & Unwarped & Warped & Unwarped & Warped \\
\hline
modis:1 & 0.7928 & \underline{\textbf{0.8207}} & 0.7942 & 0.8193 \\
modis:2 & 0.7697 & 0.7933 & 0.7813 & \textbf{0.8060} \\
modis:4 & 0.7378 & \textbf{0.7541} & 0.7202 & 0.7490 \\
modis:3 & 0.7211 & \textbf{0.7287} & 0.7197 & 0.7115 \\
modis:18 & 0.6663 & 0.6788 & \textbf{0.6800} & 0.6793 \\
modis:5 & 0.6586 & 0.6744 & 0.6734 & \textbf{0.6805} \\
modis:19 & 0.6766 & \textbf{0.6890} & 0.6430 & 0.6402 \\
modis:22 & 0.6422 & \textbf{0.6627} & 0.6304 & 0.6509 \\
modis:31 & 0.6306 & \textbf{0.6560} & 0.6324 & 0.6466 \\
modis:23 & 0.6436 & \textbf{0.6522} & 0.6252 & 0.6496 \\
modis:17 & 0.6799 & \textbf{0.6857} & 0.6044 & 0.6092 \\
modis:32 & 0.6306 & \textbf{0.6507} & 0.6246 & 0.6423 \\
modis:29 & 0.6296 & 0.6430 & 0.6236 & \textbf{0.6470} \\
modis:20 & 0.6174 & 0.6408 & 0.6351 & \textbf{0.6465} \\
modis:6 & 0.6131 & 0.6265 & 0.6173 & \textbf{0.6454} \\
\hline
\end{tabular}

%% file: model_results_table.tex
\begin{tabular}{|l|c|c|c|c|}
\hline
  & \ac{bacc} $\uparrow$ & \ac{bacc} $\uparrow$ & Macro $F_1$ $\uparrow$ & Macro $F_1$ $\uparrow$ \\
Model & Unwarped &  Warped & Unwarped &  Warped \\
\hline
\ac{lsvm} & 0.8570 & \underline{\textbf{0.8814}} & 0.8122 & \textbf{0.8362} \\
\ac{gb} & 0.8538 & \textbf{0.8658} & 0.8253 & \underline{\textbf{0.8367}} \\
Random Forest & 0.8115 & \textbf{0.8428} & 0.7633 & \textbf{0.8094} \\
RBF SVM & \textbf{0.7907} & 0.7873 & \textbf{0.8168} & 0.8122 \\
Logistic Regression & 0.7457 & \textbf{0.7740} & 0.7718 & \textbf{0.8010} \\
\hline
All Ice Baseline & \multicolumn{2}{c|}{0.5000} & \multicolumn{2}{c|}{0.2009} \\
Oracle Baseline & \multicolumn{2}{c|}{0.9100} & \multicolumn{2}{c|}{0.8993} \\
\hline
\end{tabular}

%% file: figs/egs1.pdf_tex
%% Creator: Inkscape 1.2.2 (b0a8486541, 2022-12-01), www.inkscape.org
%% Accompanies image file 'egs1.pdf' (pdf, eps, ps)
\begingroup%
  \makeatletter%
  \providecommand\color[2][]{%
    \errmessage{(Inkscape) Color is used for the text in Inkscape, but the package 'color.sty' is not loaded}%
    \renewcommand\color[2][]{}%
  }%
  \providecommand\transparent[1]{%
    \errmessage{(Inkscape) Transparency is used (non-zero) for the text in Inkscape, but the package 'transparent.sty' is not loaded}%
    \renewcommand\transparent[1]{}%
  }%
  \providecommand\rotatebox[2]{#2}%
  \newcommand*\fsize{\dimexpr\f@size pt\relax}%
  \newcommand*\lineheight[1]{\fontsize{\fsize}{#1\fsize}\selectfont}%
  \ifx\svgwidth\undefined%
    \setlength{\unitlength}{595.27559055bp}%
    \ifx\svgscale\undefined%
      \relax%
    \else%
      \setlength{\unitlength}{\unitlength * \real{\svgscale}}%
    \fi%
  \else%
    \setlength{\unitlength}{\svgwidth}%
  \fi%
  \global\let\svgwidth\undefined%
  \global\let\svgscale\undefined%
  \makeatother%
  \begin{picture}(1,1.41428571)%
    \lineheight{1}%
    \setlength\tabcolsep{0pt}%
    \put(0.13586739,0.13064298){\rotatebox{135}{\makebox(0,0)[lt]{\lineheight{1.25}\smash{\begin{tabular}[t]{l}Input RGB\end{tabular}}}}}%
    \put(0.13579309,0.25752654){\rotatebox{135}{\makebox(0,0)[lt]{\lineheight{1.25}\smash{\begin{tabular}[t]{l}Warp Map\end{tabular}}}}}%
    \put(0.13603524,0.38473754){\rotatebox{135}{\makebox(0,0)[lt]{\lineheight{1.25}\smash{\begin{tabular}[t]{l}M:1 LSVM (Unwarped)\end{tabular}}}}}%
    \put(0.13603527,0.50745532){\rotatebox{135}{\makebox(0,0)[lt]{\lineheight{1.25}\smash{\begin{tabular}[t]{l}M:1 LSVM (Warped)\end{tabular}}}}}%
    \put(0.13603527,0.63440475){\rotatebox{135}{\makebox(0,0)[lt]{\lineheight{1.25}\smash{\begin{tabular}[t]{l}LSVM (Unwarped)\end{tabular}}}}}%
    \put(0.13603524,0.75712257){\rotatebox{135}{\makebox(0,0)[lt]{\lineheight{1.25}\smash{\begin{tabular}[t]{l}LSVM (Warped)\end{tabular}}}}}%
    \put(0.13592754,0.88402734){\rotatebox{135}{\makebox(0,0)[lt]{\lineheight{1.25}\smash{\begin{tabular}[t]{l}GB (Unwarped)\end{tabular}}}}}%
    \put(0.13592755,1.00674516){\rotatebox{135}{\makebox(0,0)[lt]{\lineheight{1.25}\smash{\begin{tabular}[t]{l}GB (Warped)\end{tabular}}}}}%
    \put(0.13592754,1.1336946){\rotatebox{135}{\makebox(0,0)[lt]{\lineheight{1.25}\smash{\begin{tabular}[t]{l}CNN (Unwarped)\end{tabular}}}}}%
    \put(0.13592754,1.2564124){\rotatebox{135}{\makebox(0,0)[lt]{\lineheight{1.25}\smash{\begin{tabular}[t]{l}CNN (Warped)\end{tabular}}}}}%
    \put(0,0){\includegraphics[width=\unitlength,page=1]{egs1.pdf}}%
  \end{picture}%
\endgroup%

%% file: figs/egs2.pdf_tex
%% Creator: Inkscape 1.2.2 (b0a8486541, 2022-12-01), www.inkscape.org
%% Accompanies image file 'egs2.pdf' (pdf, eps, ps)
\begingroup%
  \makeatletter%
  \providecommand\color[2][]{%
    \errmessage{(Inkscape) Color is used for the text in Inkscape, but the package 'color.sty' is not loaded}%
    \renewcommand\color[2][]{}%
  }%
  \providecommand\transparent[1]{%
    \errmessage{(Inkscape) Transparency is used (non-zero) for the text in Inkscape, but the package 'transparent.sty' is not loaded}%
    \renewcommand\transparent[1]{}%
  }%
  \providecommand\rotatebox[2]{#2}%
  \newcommand*\fsize{\dimexpr\f@size pt\relax}%
  \newcommand*\lineheight[1]{\fontsize{\fsize}{#1\fsize}\selectfont}%
  \ifx\svgwidth\undefined%
    \setlength{\unitlength}{595.27559055bp}%
    \ifx\svgscale\undefined%
      \relax%
    \else%
      \setlength{\unitlength}{\unitlength * \real{\svgscale}}%
    \fi%
  \else%
    \setlength{\unitlength}{\svgwidth}%
  \fi%
  \global\let\svgwidth\undefined%
  \global\let\svgscale\undefined%
  \makeatother%
  \begin{picture}(1,1.41428571)%
    \lineheight{1}%
    \setlength\tabcolsep{0pt}%
    \put(0.13586739,0.13064298){\rotatebox{135}{\makebox(0,0)[lt]{\lineheight{1.25}\smash{\begin{tabular}[t]{l}Input RGB\end{tabular}}}}}%
    \put(0.13579309,0.25752654){\rotatebox{135}{\makebox(0,0)[lt]{\lineheight{1.25}\smash{\begin{tabular}[t]{l}Warp Map\end{tabular}}}}}%
    \put(0.13603524,0.38473754){\rotatebox{135}{\makebox(0,0)[lt]{\lineheight{1.25}\smash{\begin{tabular}[t]{l}M:1 LSVM (Unwarped)\end{tabular}}}}}%
    \put(0.13603527,0.50745532){\rotatebox{135}{\makebox(0,0)[lt]{\lineheight{1.25}\smash{\begin{tabular}[t]{l}M:1 LSVM (Warped)\end{tabular}}}}}%
    \put(0.13603527,0.63440475){\rotatebox{135}{\makebox(0,0)[lt]{\lineheight{1.25}\smash{\begin{tabular}[t]{l}LSVM (Unwarped)\end{tabular}}}}}%
    \put(0.13603524,0.75712257){\rotatebox{135}{\makebox(0,0)[lt]{\lineheight{1.25}\smash{\begin{tabular}[t]{l}LSVM (Warped)\end{tabular}}}}}%
    \put(0.13592754,0.88402734){\rotatebox{135}{\makebox(0,0)[lt]{\lineheight{1.25}\smash{\begin{tabular}[t]{l}GB (Unwarped)\end{tabular}}}}}%
    \put(0.13592755,1.00674516){\rotatebox{135}{\makebox(0,0)[lt]{\lineheight{1.25}\smash{\begin{tabular}[t]{l}GB (Warped)\end{tabular}}}}}%
    \put(0.13592754,1.1336946){\rotatebox{135}{\makebox(0,0)[lt]{\lineheight{1.25}\smash{\begin{tabular}[t]{l}CNN (Unwarped)\end{tabular}}}}}%
    \put(0.13592754,1.2564124){\rotatebox{135}{\makebox(0,0)[lt]{\lineheight{1.25}\smash{\begin{tabular}[t]{l}CNN (Warped)\end{tabular}}}}}%
    \put(0,0){\includegraphics[width=\unitlength,page=1]{egs2.pdf}}%
  \end{picture}%
\endgroup%

%% file: conclusion.tex
\section{Acknowledgments}
We thank David Wyld, Penelope Wagner, Andrew Fleming, and Andreas Cziferszky for supporting software, providing advice on analyzing the \ac{eo} datasets, and creating the expert sea ice pin classifications. This work was funded by NERC award NE/Z504269/1 and EPSRC award UKRI2703.

\newpage
\section{Discussion and Conclusion}

Our results demonstrate that explicitly modeling spatial and temporal correspondence between multimodal satellite image acquisitions is an effective step for improving the performance of models on downstream reasoning tasks in dynamic environments. By performing spatial evidence localization and warping imagery prior to multimodal fusion, segmentation performance of pixel-wise and deep learning classifiers significantly improved for our challenging sea ice use case, achieving a maximum \ac{bacc} score of 0.88, approaching the oracle performance of 0.91. Baseline segmentation accuracy is higher when more channels are introduced, reducing the relative gain provided by our method, although consistent improvements remain, suggesting that accurate grounding of multimodal imagery provides additional complementary information to multimodal reasoning rather than replacing it.

Our findings challenge the widespread assumption within \acp{gfm} that perceptual observations from different sensors natively correspond to the same grounded physical feature. This implicit assumption does not hold in many dynamic systems, including coastal and ocean surfaces, atmospheric systems, and natural hazards. As satellite imagery continues to increase in spatial resolution, progressively smaller physical displacements become observable. Consequently, assumptions of pixel-level correspondence between acquisitions are likely to become increasingly invalid. We hope that explicit perceptual grounding and spatial co-registration of multimodal imagery will become an increasingly important component of \ac{gfm} pretraining and that this is a first step in reducing the reliance of \acp{gfm} on multi-day mosaics of satellite data.

We also demonstrate that sparse, pixel-wise, high-quality labels can be sufficient for training dense segmentation models. Both pixel-wise and convolutional classifiers achieved comparable performance (\ac{bacc}=0.88) when trained on warped imagery, suggesting that explicitly correcting geometric misalignment is more beneficial than relying on convolutional architectures to implicitly learn spatial correspondence. This finding is consistent with the growing trend towards pixel-level representation learning in \acp{gfm} \cite{feng2026tessera, lisaius2024barlow}, suggesting that sparse, high-quality annotations may provide a scalable alternative to exhaustively labeled datasets.

Future work will investigate integrating motion-aware multimodal correspondence into \ac{gfm} pretraining, by jointly optimizing dense correspondence and multimodal representations, rather than treating it as a separate preprocessing step. We will also initialize models using other externally estimated motion fields or physics-informed priors to increase model robustness, particularly in areas of large-scale, heterogeneous image deformation. We believe this work will motivate a shift towards explicit perceptual grounding and spatial co-registration as a fundamental component of future \ac{gfm} pretraining.